\documentclass[twoside]{article}

\usepackage[accepted]{aistats2021}
\usepackage{graphicx}
\usepackage{comment}
\usepackage{amsmath,amssymb}
\usepackage{color}
\usepackage{adjustbox}
\usepackage{natbib}

\usepackage{algorithm}
\usepackage{algorithmic}
\floatstyle{ruled}
\newfloat{algorithm}{tbp}{loa}
\algsetup{linenodelimiter=}

\DeclareMathOperator*{\argmax}{arg\,max}

\DeclareMathOperator*{\st}{\text{subject to}}

\newcommand{\din}{\mathcal D}

\begin{document}

%

%

\twocolumn[

\aistatstitle{RAILS: A Robust Adversarial Immune-inspired Learning 
System}


\aistatsauthor{Ren Wang, Tianqi Chen,  Stephen Lindsly, Alnawaz Rehemtulla, Alfred Hero, Indika Rajapakse}

\aistatsaddress{University of Michigan} ]

\begin{abstract}
Adversarial attacks against deep neural networks are continuously evolving. Without effective defenses, they can lead to catastrophic failure. 
The long-standing and arguably most powerful natural defense system is 
the  mammalian immune system, which has successfully defended
against attacks by novel pathogens for millions of years. In this paper,
we propose a new adversarial defense framework, called the Robust Adversarial Immune-inspired Learning System (RAILS). RAILS incorporates an Adaptive Immune System Emulation (AISE), which emulates {\em in silico} the biological mechanisms that are used to defend the host against attacks by pathogens. 
We use RAILS to harden Deep k-Nearest Neighbor (DkNN) architectures  against evasion attacks. 
Evolutionary programming is used to simulate processes in the natural immune system: B-cell flocking, clonal expansion, and affinity maturation. We show that the RAILS learning curve exhibits similar diversity-selection learning phases as observed in our {\em in vitro} biological experiments.
When applied to adversarial image classification on three different datasets, RAILS delivers an additional $5.62\%/12.56\%/4.74\%$ robustness improvement as compared to applying DkNN alone, without appreciable loss of accuracy on clean data.

\end{abstract}

\section{INTRODUCTION}
The state-of-the-art in supervised deep learning  \citep*{lecun2015deep}, has dramatically improved over the past decade. Deep learning techniques have led to significant improvements in 
applications such as: face recognition \citep*{mehdipour2016comprehensive}; object detection \citep*{zhao2019object}; and natural language processing \citep*{young2018recent}. Despite these successes, deep learning techniques are not resilient to adversarial attacks such as  evasion attacks on the inputs and poisoning attacks on the training data \citep*{goodfellow2014explaining,szegedy2013intriguing,GDG17}. The adversarial vulnerability of deep neural networks (DNN) has restricted its applications, motivating researchers to develop effective defense methods.

Current defense methods can be broadly divided into three categories: (1) Adversarial example detection \citep*{metzen2017detecting,feinman2017detecting,grosse2017statistical,xu2017feature}; (2) Robust training \citep*{madry17,zhang2019theoretically,cohen2019certified,shafahi2019adversarial,Wong2020Fast}; and (3) Depth classifiers with natural robustness \citep*{papernot2018deep,samangouei2018defense}. The first category of methods defends the deep learning algorithm using simple models for detecting adversarial examples as outliers. However, it has been shown that adversarial detection methods are not perfect and can be easily defeated \citep*{carlini2017adversarial}. 
Robust training aims to harden the model to deactivate the evasion attack. Known robust training methods are tailored to a certain level of attack strength in the context of $\ell_p$-perturbation. Moreover, the trade-off between accuracy and robustness presents challenges \citep*{zhang2019theoretically}. Recently alternative defense strategies have been proposed that implement depth classifiers which are naturally resilient to evasion attacks. Despite these advances, current methods have difficulty providing an acceptable level of robustness to novel attacks \citep*{athalye2018obfuscated}.

Relative to designing effective defenses to reduce a system's vulnerability to attacks, a natural question to ask is the following. Can we emulate the naturally robust biological immune system to improve resiliency? 
The mammalian immune system has evolved over millions of years, resulting in a life-long learning system that continuously learns from experience to defend against a constant onslaught of diverse attacks from bacterial, viral, fungal, and other infections.  
The natural immune system has evolved a built-in detector to distinguish non-self components from self components \citep*{farmer1986immune}, and has a naturally robust architecture \citep*{mesin2016germinal}. 
Furthermore, the immune system continuously increases its degree of robustness by adaptively learning from attacks \citep*{mesin2020restricted}.

Motivated by the natural immune system's powerful evolved defense capacities,
we propose a new framework, Robust Adversarial Immune-inspired Learning System (RAILS), that can effectively defend deep learning architectures against attacks. While RAILS can be applied to defending against various attacks on the training/test data, in this paper we restrict attention to evasion attacks on the input.   

\paragraph{Contributions.}

Compared to existing defense methods, we make the following contributions:

$\bullet$ We propose a new adversarial defense framework (RAILS) that is inspired by the natural immune system and show that the learning patterns exhibited by RAILS aligns with those of the immune system.

$\bullet$ 
On three different datasets, the RAILS implementation achieves
$5.62\%/12.56\%/4.74\%$ robustness improvement over Deep k-Nearest Neighbors (DkNN) \citep*{papernot2018deep} and higher confidence scores.

$\bullet$ RAILS robustifies deep learning classifiers using an  Adaptive Immune System Emulation (AISE) that emulates adaptive learning (life-long learning) in the natural immune system by adding virtual B-cell's (memory data) to the training data. We show that AISE hardening of a DkNN provides a $2.3\%$ robustness improvement to the DkNN with only $5\%$ augmentation of the training data.

$\bullet$ The AISE emulated immune system defends against attacks via mutation and cross-over mechanisms, thus it is not restricted to $\ell_p$ or any specific type of attack.








\paragraph{Related Work.}
After it was established that DNNs were vulnerable to evasion attacks \citep*{szegedy2013intriguing}, different types of defense mechanisms have been proposed in recent years. One intuitive idea is to eliminate the adversarial examples through outlier detection. The authors of \citet*{metzen2017detecting,grosse2017statistical} considered training an additional sub-network to separate adversarial attacks from benign inputs. 
The authors of \citet*{feinman2017detecting,xu2017feature} 
use kernel density estimation and Bayesian uncertainty estimation to identify adversarial examples.
The above approaches rely on the fundamental assumption that the distributions of benign and adversarial examples are distinct, an assumption that has been challenged in \citet*{carlini2017adversarial}.

In addition to adversarial attack detection, other methods have been proposed that focus on robust training, aiming to robustify the deep architecture during the learning phase. 
Examples include projected gradient descent (PGD)-based adversarial training \citep*{madry17} and its many variants \citep*{shafahi2019adversarial,Wong2020Fast}, training via randomized smoothing with robustness guarantees within a $\ell_2$ ball \citep*{cohen2019certified}, and TRADES which optimizes the standard-robust error gap \citep*{zhang2019theoretically}.
Though these defenses are effective against adversarial examples with a certain level of $\ell_p$ attack strength 
there is a sacrifice in overall classification accuracy. In contrast, RAILS is developed to defend against diverse powerful attacks with little sacrifice in accuracy. Moreover, when implemented online, RAILS can adaptive increase robustness during the inference stage.

Another approach is to leverage on existing work on robustifying deep classifiers. An example is the deep k-Nearest Neighbor (DkNN) classifier \citep*{papernot2018deep} that robustifies against instance perturbations  
by applying kNN's to features extracted from each layer. The prediction confidence of the DkNN can be low due when there are many hidden layers. Another relevant work is the auxiliary Generative Adversarial Network (GAN) which can be used for defending against attacks \citep*{samangouei2018defense}. However, 
it is difficult to 
properly train and select hyperparameters for the GAN model.
We will show in this paper that the proposed RAILS method is an alternative that can provide improved robustness and confidence scores.

Another line of research relevant to ours is adversarial transfer learning \citep*{liu2019transferable,Shafahi2020Adversarially}, which aims to maintain robustness when there is covariate shift from training data to test data.    
We remark that covariate shift occurs in the immune system as it adapts to novel mutated pathogen strain.  

\begin{figure*}[h]
\centerline{\includegraphics[width=.9\textwidth]{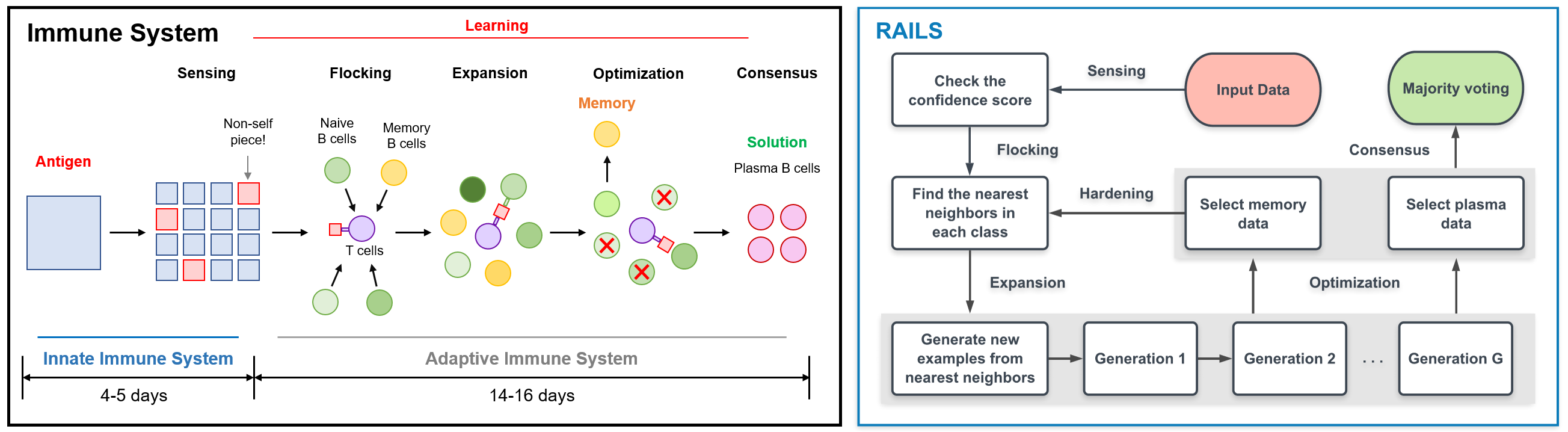}}
\caption{Simplified Immune System (left) and RAILS Computational Workflow (right).}
\label{fig: bio_comp}
\end{figure*}

\section{FROM IMMUNE SYSTEM TO COMPUTATIONAL SYSTEM}

\subsection{Learning Strategies of Immune System}

Systems robustness is a property that must be designed into the architecture, and one of the greatest examples of this is within the mammalian adaptive immune system \citep*{rajapakse2011emerging}. The architecture of the adaptive immune system ensures a robust response to foreign antigens, splitting the work between active sensing and competitive growth to produce an effective antibody. \textit{Sensing} of a foreign attack leads to antigen-specific B-cells \textit{flocking} to lymph nodes, and forming temporary structures called germinal centers \citep*{de2015dynamics,farmer1986immune}. In the \textit{expansion} phase, a diverse initial set of B-cells bearing antigen-specific immunoglobulins divide symmetrically to populate the germinal center in preparation for and \textit{optimization} (affinity maturation). The B-cells with the highest affinity to the antigen are selected
to asymmetrically divide and mutate, which leads to new B-cells with higher affinity to the antigen \citep{mesin2016germinal}. Memory B-cells are stored within this step, which can be used to defend against similar attacks in the future. B-cells that reach \textit{consensus}, or achieve a threshold affinity against the foreign antigen, undergo terminal differentiation into plasma B-cells. Plasma B-cells represent the actuators of the humoral adaptive immune response. The adaptive immune system is incredibly complex, but we can simplify its robust learning process into these five steps: sensing, flocking, expansion, optimization, and consensus (Figure \ref{fig: bio_comp}) \citep*{cucker2007emergent,rajapakse2017emergence}.

\subsection{From Biology to Computation}
Motivated by recent advances in understanding the biological immune system, we propose a new {\em in silico} defense strategy - the  Robust Adversarial Immune-inspired Learning System (RAILS). This computational system is closely associated with  the simplified architecture of the immune system \citep{RL2020}. Figure~\ref{fig: bio_comp} displays a comparison between the immune system workflow and the RAILS workflow. Both systems are composed of a five-step process. For example, RAILS emulates clonal expansion from the immune system by enlarging the population of candidates (B-cells). Similar to the plasma B-cells and memory B-cells generated in the immune system, RAILS generates plasma data for predictions of the present inputs and generates memory data for the defense against future attacks. 

\begin{figure}[h]
\centerline{\includegraphics[width=.43\textwidth]{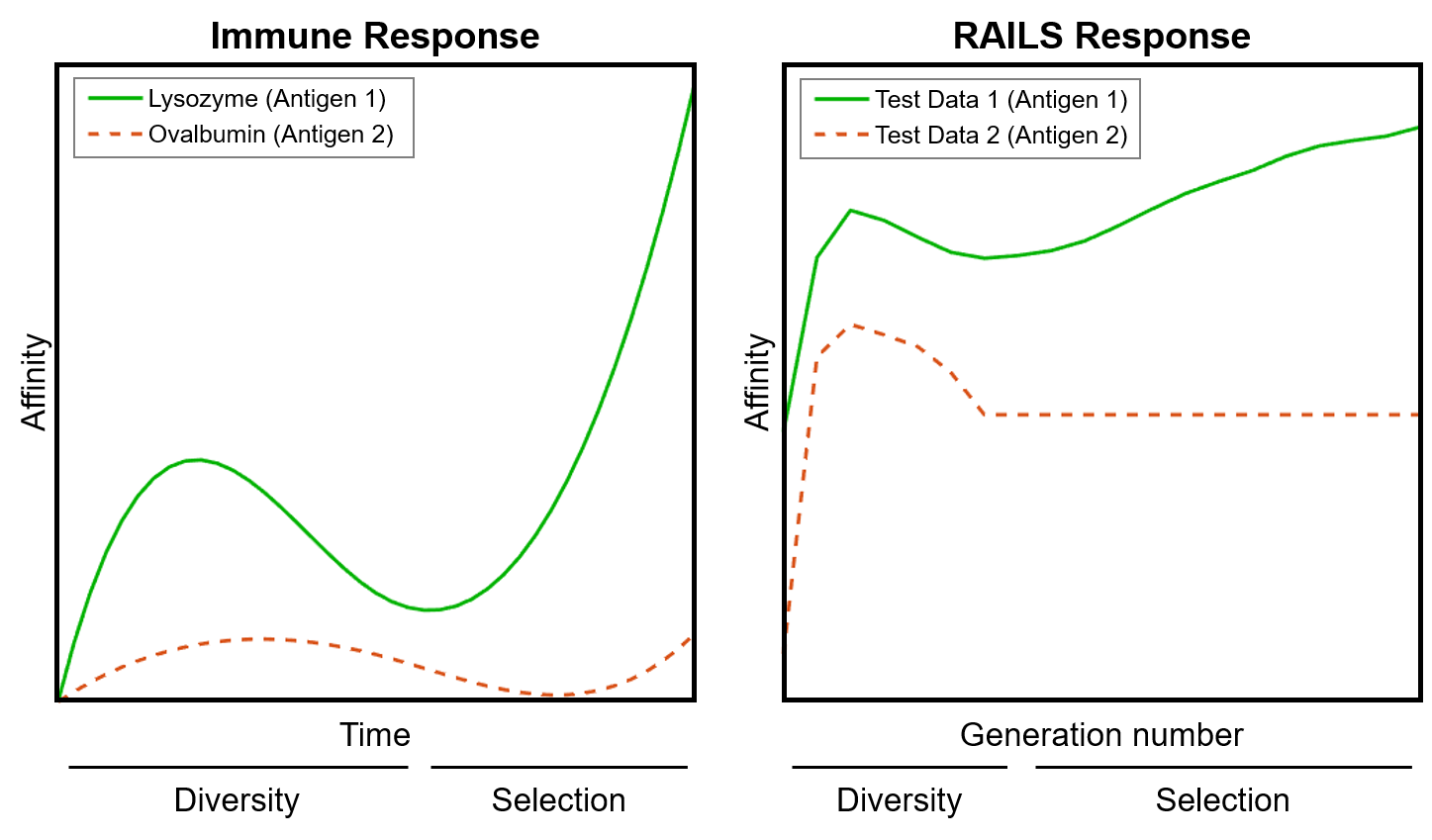}}
\caption{Correspondence of Immune System \textit{in vitro} Analog (left) and RAILS (right) Experiments}
\label{fig: curves_bio_rails}
\end{figure}

To demonstrate that the proposed RAILS computational system captures important properties of the immune system, we compare the learning curves of the two systems in Figure~\ref{fig: curves_bio_rails}. The green and red lines depict the affinity change between the population and the antigen (test data). The data selected in the flocking step comes from antigen 1 (test data 1) in all tests, which results in a low-affinity increase for antigen 2 (test data 2). One can see that both the immune system's learning curve of antigen 1 and the RAILS learning curve of antigen 1 have a small affinity decrease at the beginning and then monotonically increases (green curves). This phenomenon indicates a two-phase learning process. The diversity phase comes from the {\em clonal expansion} that generates diverse data points (virtual B-Cells), which results in {\em decreasing affinity}. The selection phase arises from selection of the high-affinity data points (virtual B-Cells) generated during the diversity phase, which results in the {\em increasing affinity}.


\section{RAILS: Overview 
}\label{sec: RAILS}

\begin{figure}[h]
\vspace{.01in}
\centerline{\includegraphics[trim=95 0 0 0,clip,width=.38\textwidth]{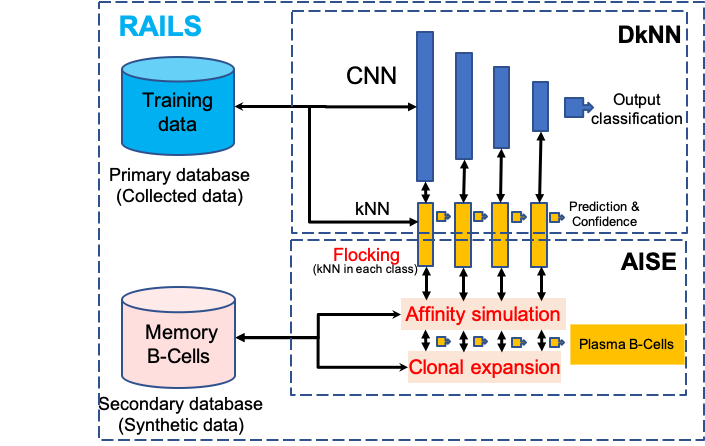}}
\vspace{-.03in}
\caption{The Architecture of RAILS}
\label{fig: arch}
\end{figure}

In this section, we first introduce the Adaptive Immune System Emulation (AISE), which is the key component of RAILS,  containing two learning stages. We then provide details of the RAILS workflow. We illustrate RAILS and AISE in the context of hardening a DkNN architecture against evasion attacks. For any input and for each hidden layer, the DkNN finds the nearest neighbors of the activation responses to the input relative to the activation responses to the training samples. Different from the original DkNN, in RAILS the k-nearest neighbor search is restricted to the training samples in a given class. AISE emulates the plasma B-cell and memory B-cell defense mechanisms to robustify the architecture. The architecture of RAILS is illustrated in Figure~\ref{fig: arch}.

\subsection{Adaptive Immune System Emulation (AISE)}
The Adaptive Immune System Emulation (AISE) is designed and implemented with a bionic process inspired by the mammalian immune system. Concretely, AISE generates virtual plasma data (plasma B-cells) and memory data (memory B-cells) through multiple generations of evolutionary programming. The plasma data and memory data are selected in different ways, thus contributing to different model robustifying levels. The plasma data contributes to the robust predictions of the present inputs, and the memory data helps to adjust the classifiers to effectively defend against future attacks. From the perspective of classifier adjustment, AISE's learning stages can be divided into \textit{Static Learning} and \textit{Adaptive Learning}.

\paragraph{Defense with Static Learning.}
Static learning helps to correct the predictions of the present inputs. Recall that the DkNN integrates the predicted k-nearest neighbors of layers in the deep neural network, and the final prediction $y_{\text{DkNN}}$ can be obtained by the following formula:
\begin{align}\label{eq: dknn}
    \begin{array}{ll}
y_{\text{DkNN}} =   \argmax_c \sum_{l=1}^L p_l^c(\mathbf x)\\  \st~~~ c \in [C]
    \end{array}
\end{align}
where $l$ is the $l$-th layer of a DNN with $L$ layers in total. $p_l^c(\mathbf x)$ is the probability of class $c$ predicted by the kNN in layer $l$ for input $\mathbf x$. There is a finite set of classes and the total number is $C$. $[C]$ denotes the set $\{1,2,\cdots,C\}$. Note that $p_l^c(\mathbf x)$ could be small for poisoned data, containing an adversarial training example, even if $y_{\text{true}}$ is the true class $c$. The purpose of static learning is to increase $p_l^{y_{\text{true}}}(\mathbf x)$ 
of the present input $\mathbf x$. The key idea is to generate new training examples via clonal expansion and optimization, and only select the examples with high affinity (plasma data) to the input. Our hypothesis is that examples inherited from parents of class $y_{\text{true}}$ have higher chance of reaching high affinity and, therefore, survival. After optimization, a majority vote is used to make the class prediction. We refer readers to Section~\ref{sec: alg_detail} for more implementation details and Section~\ref{sec: perf_layer} for visualization.

\paragraph{Defense with Adaptive Learning.}
Different from static learning, adaptive learning tries to harden the classifier to defend against potential future attacks. 
The hardening is done by leveraging another set of data - memory data generated during affinity maturation. Unlike plasma data, memory data is selected from examples with moderate-affinity to the input, which can rapidly adapt to new variants of the current adversarial examples. 
Adaptive learning is a life-long learning process and will provide a naturally high $p_l^{y_{\text{true}}}(\mathbf x)$ even if using the DkNN alone. This paper will mainly focus on static learning and single-stage adaptive learning that implements a single cycle of classifier hardening.

\subsection{RAILS Details}\label{sec: alg_detail}

\subsubsection{Five-Step Workflow}

Given a mapping $F: \mathbb{R}^{d} \rightarrow \mathbb{R}^{d_{\prime}}$ and two vectors $\mathbf x_1, \mathbf x_2 \in \mathbb{R}^{d}$, we first define the affinity score between $\mathbf x_1$ and $\mathbf x_2$ as $A(F;\mathbf x_1, \mathbf x_2) = -\|F(\mathbf x_1) - F(\mathbf x_2)\|_2$, where $A$ is a affinity function equal to the negative Euclidean distance. In the DNN context, $F$ denotes the feature mapping from input to feature representation, and $A$ measures the similarity between two inputs. Higher affinity score indicates higher similarity. Algorithm~\ref{RAILS} shows the five-step workflow of RAILS: Sensing, Flocking, Expansion, Optimization, and Consensus. We explain each step below.

\paragraph{Sensing.} This step performs the initial discrimination between adversarial and clean inputs. This is implemented using an outlier detection procedure, for which there are several different methods available \citep*{feinman2017detecting,xu2017feature}. The DkNN provides a {\em credibility metric} that can measure the label consistency of k-nearest neighbors in each layer. The higher the credibility, the higher the confidence that the input is clean. The sensing stage provides a confidence score of the DkNN architecture and does not affect RAILS's predictions. 

\paragraph{Flocking.} This step provides a starting point for clonal expansion. For each class and each selected layer, we find the k-nearest neighbors that have the highest initial affinity score to the input data. Mathematically, we select
\begin{align}\label{eq: knn}
    \begin{array}{ll}
\mathcal{N}_l^c = \{(\hat{\mathbf x}, y_c)|R_c(\hat{\mathbf x}) \le k, (\hat{\mathbf x}, y_c) \in \din_c\} \\ \text{Given}\\ A(f_l; \mathbf x_i^c, \mathbf x) \le A(f_l; \mathbf x_j^c, \mathbf x) \Leftarrow R_c(i) > R_c(j)\\ \forall c \in [C], l \in \mathcal{L},  \forall i, j \in [n_c],
    \end{array}
\end{align}
where $x$ is the input. $\mathcal{L}$ is the set of the selected layers. $\din_c$ is the training dataset from class $c$ and the size $|\din_c|=n_c$. $R_c: [n_c] \rightarrow [n_c]$ is a ranking function that sorts the indices based on the affinity score. If the memory database has been populated, we will find the nearest neighbors using both the training data and the memory data generated at the previous iteration.

\paragraph{Expansion.} This step generates new examples (offspring) from the existing examples (parents). The ancestors are nearest neighbors found by Flocking. The process can be viewed as creating new nodes (offspring) linked to the existing nodes (parents), and can be analogous to 
Preferential Attachment \citep*{barabasi1999emergence}. The probability of a new node linking to node $i$ is
\begin{align}\label{eq: pre_att}
    \begin{array}{ll}
\Pi(k_i) = \frac{k_i}{\sum_j k_j},
    \end{array}
\end{align}
where $k_i$ is the degree of node $i$. In preferential attachment models new nodes prefer to attach to existing nodes with high vertex degree. In RAILS, we use a surrogate for the degree, which is the exponentiated affinity measure, and the offspring are generated by parents having high degree
The diversity generated during Expansion is provided by the operations \textit{selection}, \textit{mutation}, and \textit{cross-over}. After new examples are generated, RAILS calculates each new example’s affinity score relative to the input. The new examples are associated with labels that are inherited from their parents. We refer readers to Section~\ref{sec: dt_clon} for additional implementation details.

\paragraph{Optimization (Affinity Maturation).} In this step, RAILS selects generated examples with high-affinity scores to be plasma data, and examples with moderate-affinity scores are saved as memory data. The selection is based on a ranking function. 
\begin{align}\label{eq: aff_mat}
    \begin{array}{ll}
S_{\text{opt}} = \{(\Tilde{\mathbf x}, \Tilde{y})|R_g(\Tilde{\mathbf x}) \le \gamma|\mathcal{P}^{(G)}|, (\Tilde{\mathbf x}, \Tilde{y}) \in \mathcal{P}^{(G)}\}
    \end{array}
\end{align}
where $R_g: [|\mathcal{P}^{(G)}|] \rightarrow [|\mathcal{P}^{(G)}|]$ is the same ranking function as $R_c$ except that the domain is the set of cardinality of the final population $\mathcal{P}^{(G)}$. $\gamma$ is a percentage parameter and is selected as $0.05$ and $0.25$ for plasma data and memory data, respectively. Note that the memory data can be selected in each generation and in a nonlinear way. For simplicity, we select memory data only in last generation. Memory data will be saved in the secondary database and used for model hardening. Details are shown in Section~\ref{sec: dt_clon}.

\paragraph{Consensus.} Note that all examples are associated with a label. In this step, plasma data use majority voting for prediction of $\mathbf x$.

\begin{algorithm}[h]
\caption{Robust Adversarial Immune-inspired Learning System (RAILS)}
\label{RAILS}
\begin{algorithmic}[1]
\REQUIRE Test data point $\mathbf x$; Training dataset $\din_{tr} = \{\din_{1}, \din_{2},\cdots,\din_{C}\}$; Number of Classes $C$; Model $M$ with feature mapping $f_l(\cdot)$ in layer $l$, $l \in \mathcal{L}$; Affinity function $A$.\\
\noindent{\textbf{First Step: Sensing}}
\STATE{Check the confidence score given by the DkNN to detect the threat of $x$.}\\
\noindent{\textbf{Second Step: Flocking}}
\FOR{$c = 1, 2, \dots, C$}
\STATE{In each layer $l \in \mathcal{L}$, find the k-nearest neighbors $\mathcal{N}_l^c$ of $\mathbf x$ in $\din_{c}$ by ranking the affinity score $A(f_{l}; \mathbf x_j, \mathbf x), \mathbf x_{j} \in \din_{c}$}
\ENDFOR\\
\noindent{\textbf{Third and Fourth Steps: Expansion and Optimization}}
\STATE{Return plasma data $S_{p}$ and memory data $S_{m}$ by using subroutine: Algorithm~\ref{Clo_Exp}}\\
\noindent{\textbf{Fifth Step: Consensus}}
\STATE{Obtain the prediction $y$ of $\mathbf x$ using the majority vote of the plasma data}
\RETURN $y$, the memory data
\end{algorithmic}
\end{algorithm}

\subsubsection{Clonal Expansion and Affinity Maturation}\label{sec: dt_clon}

Clonal expansion and affinity maturation (optimization) are the two main steps that occur after flocking. Details are shown in Algorithm~\ref{Clo_Exp}. The goal is to promote diversity and explore the best solutions in a broader search space. Finally, plasma data and memory data are generated. Three operations support the creation of new examples: selection, cross-over, and mutation. We introduce them in detail below. 

\paragraph{Selection.} The selection operation aims to decide which candidates in the generation will be chosen to generate the offspring. We will calculate the probability for each candidate through a softmax function.
\begin{align}\label{eq: prob}
    \begin{array}{ll}
P(\mathbf x_i) &= Softmax(A(f_{l}; \mathbf x_i, \mathbf x)/\tau) \\&= \frac{\exp{(A(f_{l}; \mathbf x_i, \mathbf x)/\tau)}}{\sum_{\mathbf x_j \in S}\exp{(A(f_{l}; \mathbf x_j, \mathbf x)/\tau)}}
    \end{array}
\end{align}
where $S$ is the set containing data points and $x_i \in S$. $\tau>0$ is the sampling temperature that controls the distance after softmax operation. Given the probability $P$ of a candidates set $S$ that is calculated through \eqref{eq: prob}, the selection operation is to randomly pick one example pair $(\mathbf x_i, y_i)$ from $S$ according to its probability.
\begin{align}\label{eq: selection}
    \begin{array}{ll}
(\mathbf x_i, y_i) = Selection(S, P)
    \end{array}
\end{align}

In RAILS, we select two parents for each offspring, and the second parent is selected from the same class as that of the first parent. The parent selection process appears in lines 6 - 11 of Algorithm~\ref{Clo_Exp}.

\paragraph{Cross-over.} The crossover operator combines different candidates (parents) for generating new examples (offspring). Given two parents $\mathbf x_{p}$ and $\mathbf x_{p}^{\prime}$, the new offspring are generated by selecting each entry (e.g., pixel) from either $\mathbf x_{p}$ or $\mathbf x_{p}^{\prime}$ via calculating the corresponding probability. Mathematically,
\begin{align}\label{eq: crossover}
    \begin{array}{ll}
\mathbf x_{\text{os}}^{\prime} = Crossover(\mathbf x_{p}, \mathbf x_{p}^{\prime}) =\\ \left \{ \begin{array}{rcl}
\mathbf x_{p}^{(i)} & \mbox{with prob}~ \frac{A(f_{l}; \mathbf x_{p}, \mathbf x)}{A(f_{l}; \mathbf x_{p}, \mathbf x)+A(f_{l}; \mathbf x_{p}^{\prime}, \mathbf x)} \\ \mathbf x_{p}^{\prime{(i)}} & \text{with prob}~ \frac{A(f_{l}; \mathbf x_{p}^{\prime}, \mathbf x)}{A(f_{l}; \mathbf x_{p}, \mathbf x)+A(f_{l}; \mathbf x_{p}^{\prime}, \mathbf x)}
\end{array} \right. \forall i \in [d]
    \end{array}
\end{align}
where $i$ represents the $i$-th entry of the example and $d$ is the dimension of the example. The cross-over operator appears in line 12 of Algorithm~\ref{Clo_Exp}.

\paragraph{Mutation.} This operation mutates each entry with probability $\rho$ by adding uniformly distributed noise in the range $[-\delta_{\text{max}},-\delta_{\text{min}}] \cup [\delta_{\text{min}},\delta_{\text{max}}]$. The resulting perturbation vector is subsequently clipped to satisfy the domain constraints.
\begin{align}\label{eq: mutation}
    \begin{array}{ll}
 \mathbf x_{\text{os}} = Mutation(\mathbf x_{\text{os}}^{\prime}) = \text{Clip}_{[0,1]}\big(\mathbf x_{\text{os}}^{\prime} +\\ \boldsymbol{1}_{[Bernoulli(\rho)]}  \mathbf u([-\delta_{\text{max}},-\delta_{\text{min}}] \cup [\delta_{\text{min}},\delta_{\text{max}}])\big)
    \end{array}
\end{align}
where $\boldsymbol{1}_{[Bernoulli(\rho)]}$ takes value $1$ with probability $\rho$ and value $0$ with probability $1-\rho$. $\mathbf u([-\delta_{\text{max}},-\delta_{\text{min}}] \cup [\delta_{\text{min}},\delta_{\text{max}}])$ is the vector that each entry is i.i.d. chosen from the uniform distribution $\mathcal{U}([-\delta_{\text{max}},-\delta_{\text{min}}] \cup [\delta_{\text{min}},\delta_{\text{max}}])$. $\text{Clip}_{[0,1]}(\mathbf x)$ is equivalent to $\max(0, \min(\mathbf x, 1))$. The mutation operation appears in line 3 and line 16 in Algorithm~\ref{Clo_Exp}.

\begin{algorithm}[h]
\caption{clonal Expansion \& Affinity Maturation}
\label{Clo_Exp}
\begin{algorithmic}[1]
\REQUIRE $\mathbf x$; K-nearest neighbors in each layer $\mathcal{N}_l^c, c \in [C], l \in \mathcal{L}$; Number of population $T$;\;Maximum generation number $G$;\;Mutation probability $\rho$;\;Mutation range parameters $\delta_{\text{min}}, \delta_{\text{max}}$;\;Sampling temperature $\tau$

\STATE{\textbf{For} each layer $l \in \mathcal{L}$, \textbf{do}}
\FORALL{$\mathbf x^{\prime} \in \textit{Union}(\{\mathcal{N}_l^1,\mathcal{N}_l^2,\cdots,\mathcal{N}_l^C\})$}
\STATE{$\mathcal{P}^{(0)} \longleftarrow Mutation(\mathbf x^{\prime})$ for $P/CK$ times}
\ENDFOR
\FOR{$g = 1, 2, \dots, G$ }
\STATE{$P_{g-1}$ = \textit{Softmax}($A(f_{l}; \mathcal{P}^{(g-1)}, \mathbf x)/\tau$)}
\FOR{$t = 1, 2, \dots, T$}
\STATE{$(\mathbf x_p, y_p) = Selection(P_{g-1}, \mathcal{P}^{(g-1)})$}
\STATE{Pick all the data $S_{y_p}$ belonging to class $y_p$ in $\mathcal{P}^{(g-1)}/\{\mathbf x_p\}$ and calculate the probability $P_{y_p} = \frac{P_{g-1}(S_{y_p})}{\sum P_{g-1}(S_{y_p})}$}
\IF{$S_{y_p} \neq \emptyset$}
\STATE{$(\mathbf x_{p}^{\prime}, y_p) = Selection(P_{y_p}, S_{y_p})$}
\STATE{$\mathbf x_{\text{os}}^{\prime} = Crossover(\mathbf x_p,\mathbf x_{p}^{\prime})$}
\ELSE
\STATE{$\mathbf x_{\text{os}}^{\prime} = \mathbf x_p$}
\ENDIF
\STATE{$\mathbf x_{\text{os}} = Mutation(\mathbf x_{\text{os}}^{\prime})$ }
\STATE{$\mathcal{P}_{p}^{(g)} \longleftarrow \mathbf x_{\text{os}}$}
\ENDFOR
\ENDFOR

\STATE{Calculate the affinity score $A(f_{l}; \mathcal{P}^{(G)}, \mathbf x), \forall l \in \mathcal{L}$}
\STATE{Select the top $5\%$ as plasma data $S_{p}^l$ and the top $25\%$ as memory data $S_{m}^l$ based on the affinity scores, $\forall l \in \mathcal{L}$}
\STATE{\textbf{end For}}
\RETURN $S_{p} = \{S_{p}^1, S_{p}^2, \cdots, S_{p}^{|\mathcal{L}|}\}$ and $S_{m} = \{S_{m}^1, S_{m}^2,\cdots,S_{m}^{|\mathcal{L}|}\}$
\end{algorithmic}
\end{algorithm}

\section{EXPERIMENTAL RESULTS}
We conducted experiments in the context of image classification. We compare RAILS to standard Convolutional Neural Network Classification (CNN) and Deep k-Nearest Neighbors Classification (DkNN) \citep*{papernot2018deep} on the MNIST \citep*{lecun1998gradient}, SVHN \citep*{netzer2011reading}, and CIFAR-10 \citep*{Krizhevsky2009learning}. We test our framework using a four-convolutional-layer neural network for MNIST, and VGG16 \citep*{SZ2014} for SVHN and CIFAR-10. We refer readers to 
Section~2 in our Supplementary for more details of datasets, models, and parameter selection. In addition to the clean test examples, we also generate the same amount of adversarial examples using a $20$($10$)-step PGD attack \citep*{madry17} for MNIST (SVHN and CIFAR-10). The attack strength is $\epsilon = 40/60$ for MNIST, and $\epsilon = 8$ for SVHN and CIFAR-10 by default. The performance will be measured by standard accuracy (SA) evaluated using benign (unperturbed) test examples and robust accuracy (RA) evaluated using the adversarial (perturbed) test examples.

\subsection{Performance in Single Layer}\label{sec: perf_layer}

\begin{figure}[h]
\vspace*{-0.05in}
  \centering
  \begin{adjustbox}{max width=0.42\textwidth }
  \begin{tabular}{@{\hskip 0.00in}c  @{\hskip 0.00in}c @{\hskip 0.02in} @{\hskip 0.02in} c @{\hskip 0.02in} @{\hskip 0.02in}c }
 \begin{tabular}{@{}c@{}}  
\vspace*{0.005in}\\
\rotatebox{90}{\parbox{17em}{\centering \Large \textbf{Adv examples ($\epsilon=60$)}}}
 \\
\end{tabular} 
&
\begin{tabular}{@{\hskip 0.02in}c@{\hskip 0.02in}}
     \begin{tabular}{@{\hskip 0.00in}c@{\hskip 0.00in}}
     \parbox{10em}{\centering \Large Conv1}  
    \end{tabular} 
    \\
 \parbox[c]{20em}{\includegraphics[width=20em]{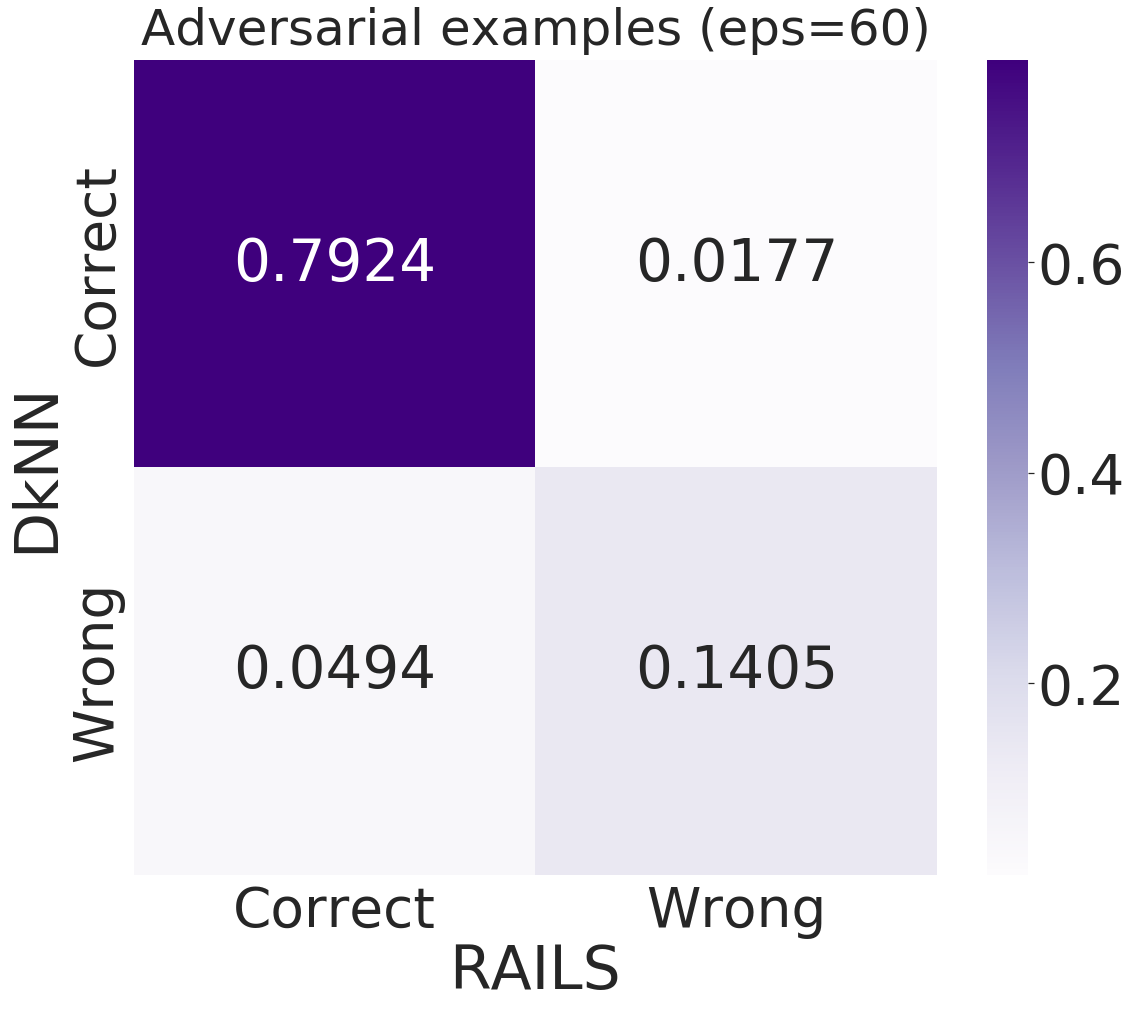}} 
\end{tabular}
&
 \begin{tabular}{@{\hskip 0.02in}c@{\hskip 0.02in}}
      \begin{tabular}{@{\hskip 0.00in}c@{\hskip 0.00in}}
     \parbox{10em}{\centering \Large Conv2}
    \end{tabular} 
    \\
 \parbox[c]{20em}{\includegraphics[width=20em]{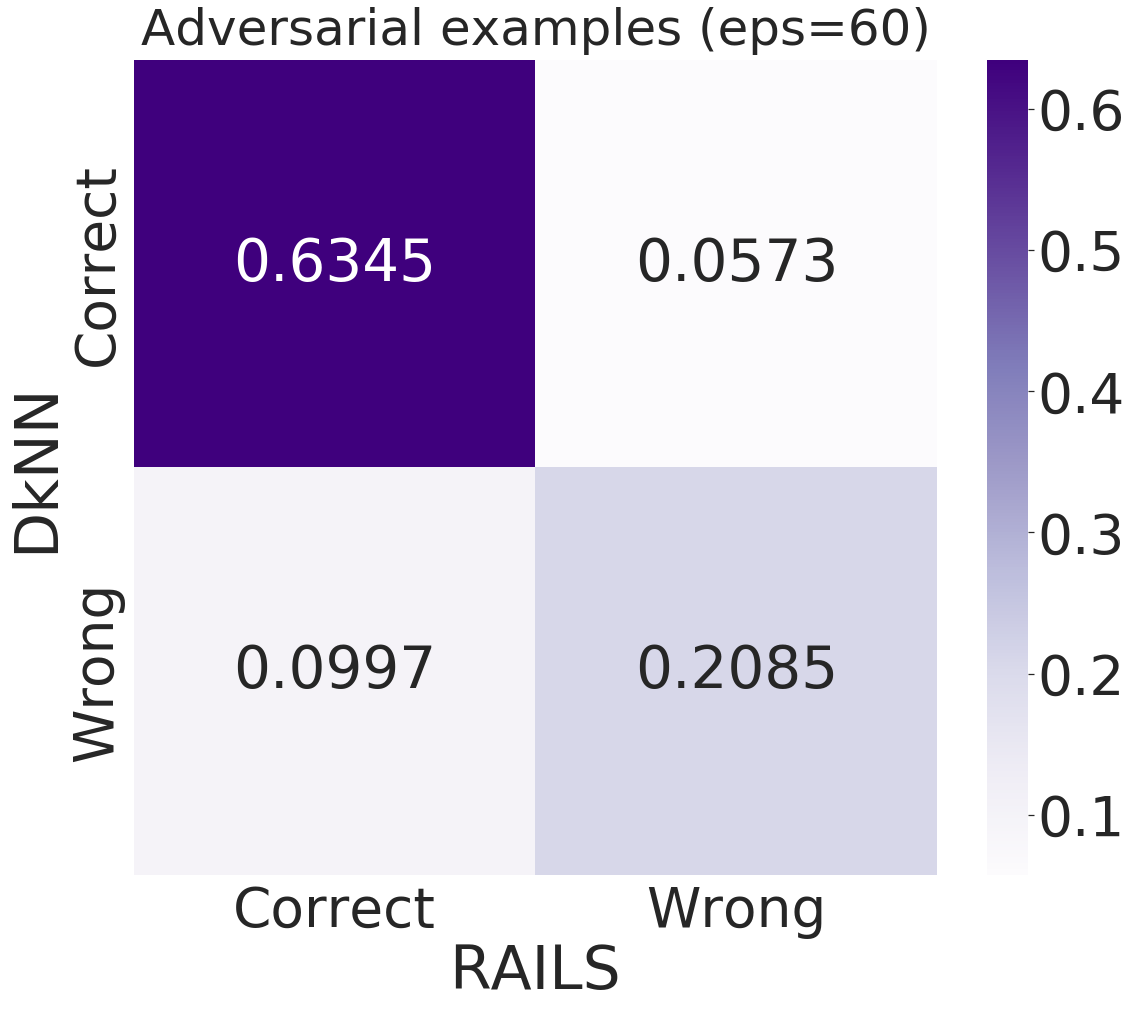}} 
\end{tabular}

\end{tabular}
  \end{adjustbox}
    \caption{Confusion Matrices of Adversarial Examples Classification in Conv1 and Conv2 (RAILS vs. DkNN)}
  \label{fig: conf_mat}
\end{figure}

\paragraph{RAILS Improves Single Layer DkNN.}
We first test RAILS in a single layer of the CNN model and compare the obtained accuracy with the results from the DkNN. Table~\ref{tab1: acc_layer} shows the comparisons in the input layer, the first convolutional layer (Conv1), and the second convolutional layer (Conv2) on MNIST. One can see that for both standard accuracy and robust accuracy, RAILS can improve DkNN in the hidden layers and achieve better results in the input layer. The input layer results indicate that RAILS can also outperform supervised learning methods like kNN. The confusion matrices in Figure~\ref{fig: conf_mat} show that RAILS has fewer incorrect predictions for those data that DkNN gets wrong. Each value in Figure~\ref{fig: conf_mat} represents the percentage of intersections of RAILS (correct or wrong) and DkNN (correct or wrong).



\begin{table}[h]
\vspace*{-0.1in}
\begin{center}
\caption{
SA/RA Performance of RAILS versus DkNN in Single Layer (MNIST)}
\label{tab1: acc_layer}
\vspace{.04in}
\resizebox{0.48\textwidth}{!}{
\begin{tabular}{lc||c|c|c}
\hline
\hline
 &  &  Input&  Conv1& Conv2 \\
\hline 
SA &  \bf{RAILS}& \bf{97.53\%} & \bf{97.77\%} & \bf{97.78\%} \\
 & DkNN & 96.88\%  & 97.4\% & 97.42\% \\
\hline
RA & \bf{RAILS}& \bf{93.78\%}  & \bf{92.56\%} & \bf{89.29\%} \\
($\epsilon=40$) & DkNN& 91.81\%  & 90.84\% & 88.26\% \\
\hline 
RA &  \bf{RAILS}& \bf{88.83\%} & \bf{84.18\%} & \bf{73.42\%} \\
($\epsilon=60$) &  DkNN& 85.54\% & 81.01\% & 69.18\% \\

\hline
\hline
\end{tabular}}
\end{center}
\end{table}


\begin{figure}[h]
\vspace*{-0.05in}
  \centering
  \begin{adjustbox}{max width=0.95\textwidth }
    \begin{tabular}{@{\hskip 0.00in}c  @{\hskip 0.02in} c @{\hskip 0.02in} @{\hskip 0.02in} c }




 \begin{tabular}{@{}c@{}}  
\rotatebox{90}{\parbox{6em}{\centering \scriptsize\textbf{\begin{tabular}[c]{@{}c@{}}  Inputs  \end{tabular}}}}
 \\

\rotatebox{90}{\parbox{7em}{\raggedleft \scriptsize \textbf{\begin{tabular}[c]{@{}c@{}}Proportion\\(true class data) \end{tabular}}}}
\\
\rotatebox{90}{\parbox{8.5em}{\raggedleft \scriptsize \textbf{\begin{tabular}[c]{@{}c@{}} Affinity avg\\(true class data)  \end{tabular} }}}
\\
\end{tabular} 
&
 \begin{tabular}{@{\hskip 0.02in}c@{\hskip 0.02in}c@{\hskip 0.02in}}
 \begin{tabular}{@{\hskip 0.02in}c@{\hskip 0.02in}c@{\hskip 0.02in}c@{\hskip 0.02in}c@{\hskip 0.02in}}
\begin{tabular}{@{\hskip 0.02in}c@{\hskip 0.02in}}
\\
 \parbox[c]{6.5em}{              }{\includegraphics[width=6em]{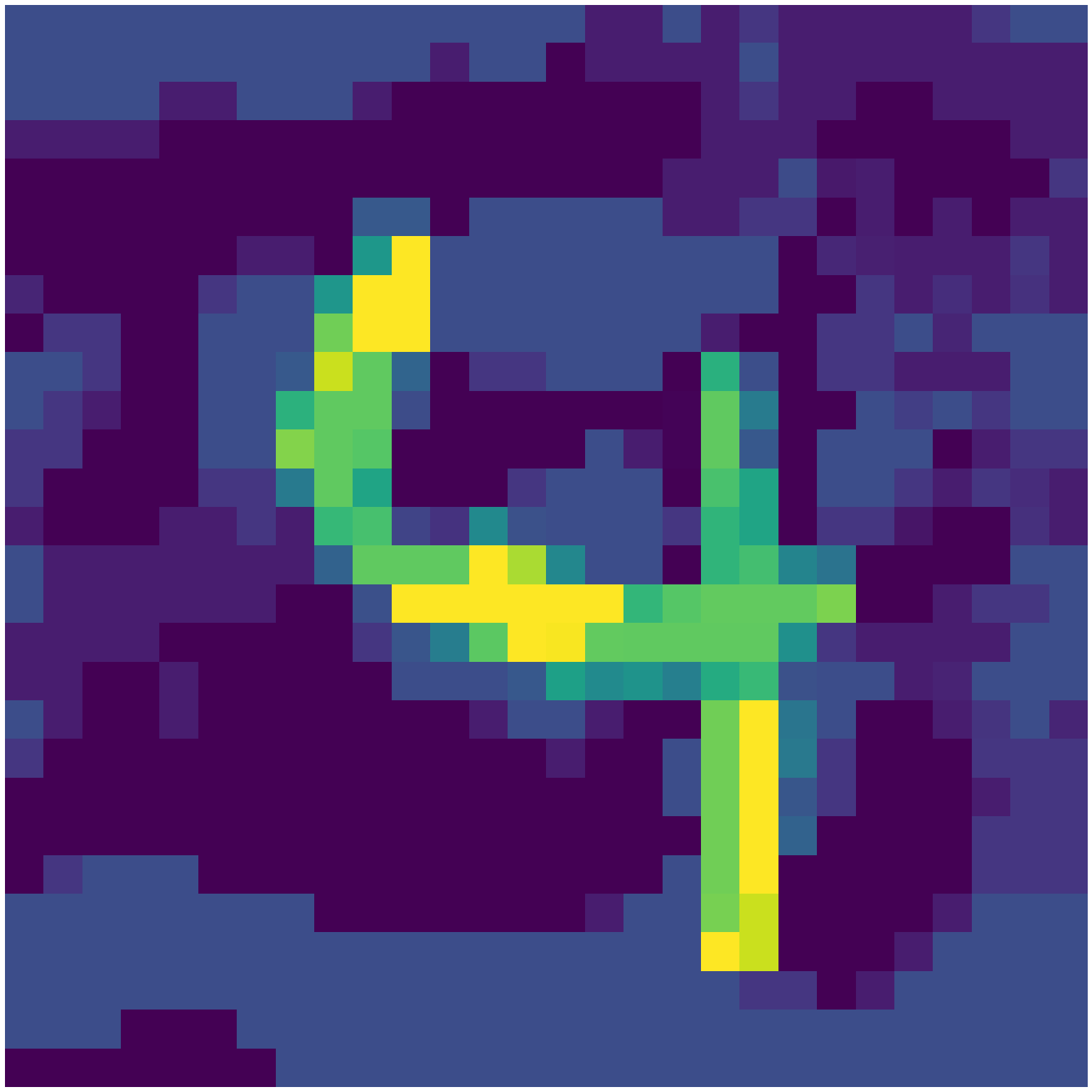}} 
  \\
  \parbox[c]{10em}{\includegraphics[trim=5 0 0 0,clip,width=9.5em]{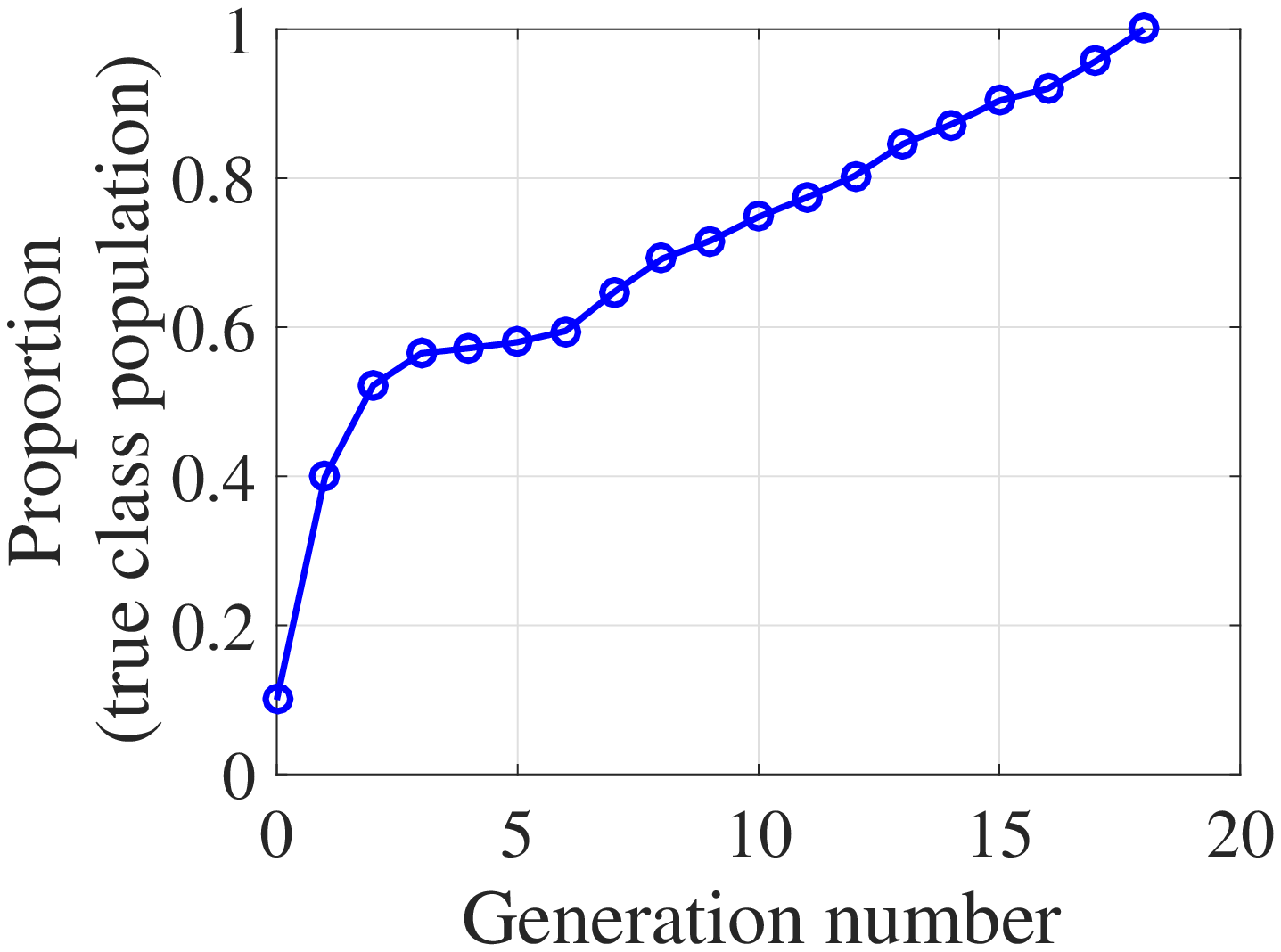}}
  \\
 \parbox[c]{10em}{\includegraphics[trim=5 0 0 0,clip,width=9.5em]{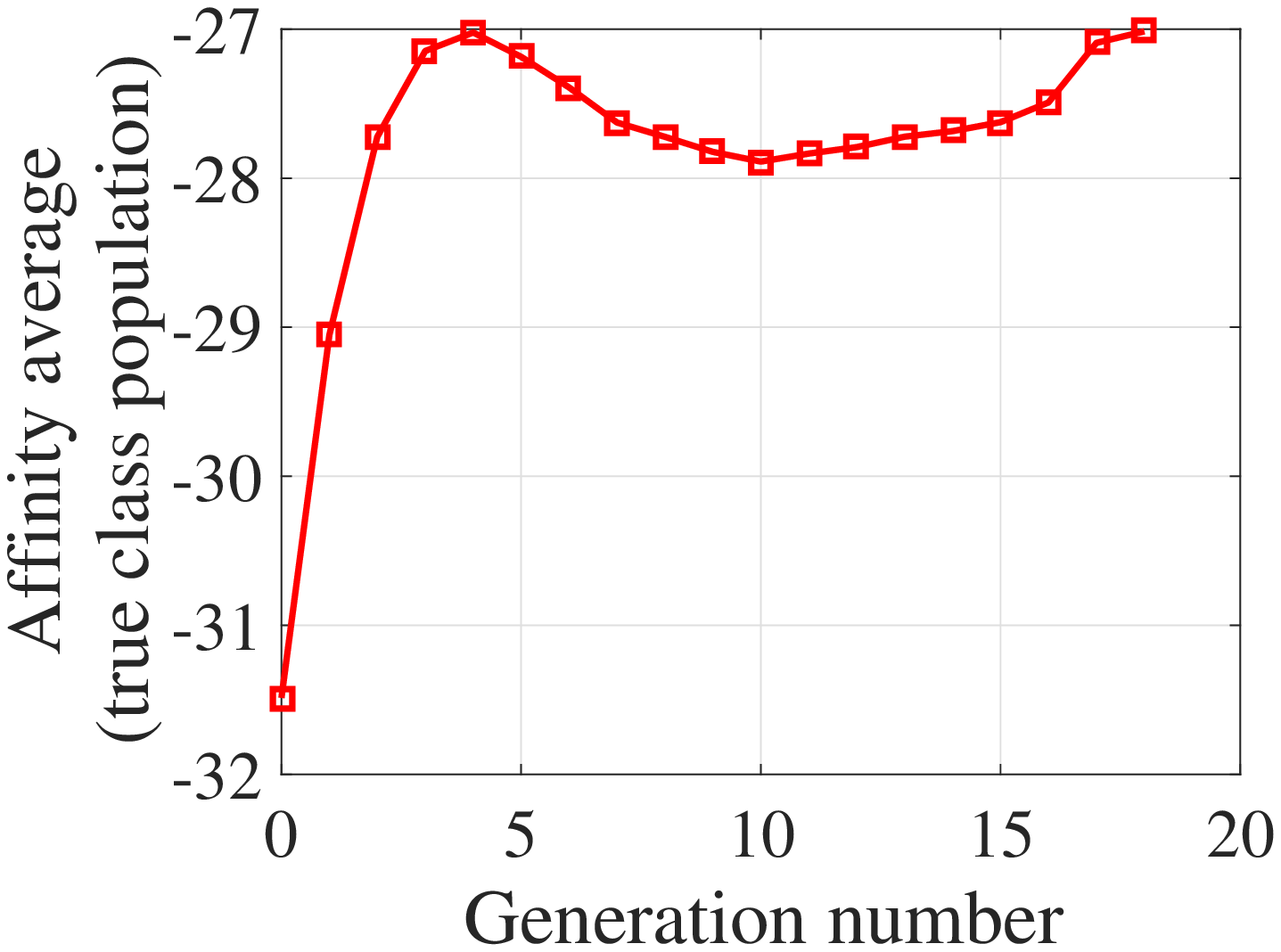}}
 \\
 \vspace{0.05in}
\begin{tabular}{@{\hskip 0.00in}c@{\hskip 0.00in}}
\parbox{9em}{\centering \scriptsize  \textbf{\begin{tabular}[c]{@{}c@{}}RAILS:4 VS DkNN:4\\RAILS confidenc: 1.0\\DkNN confidence: 0.6  \end{tabular}}} 
\end{tabular}
\end{tabular}

&
 \begin{tabular}{@{\hskip 0.02in}c@{\hskip 0.02in}}
\\
 \parbox[c]{6.5em}{              }{\includegraphics[width=6em]{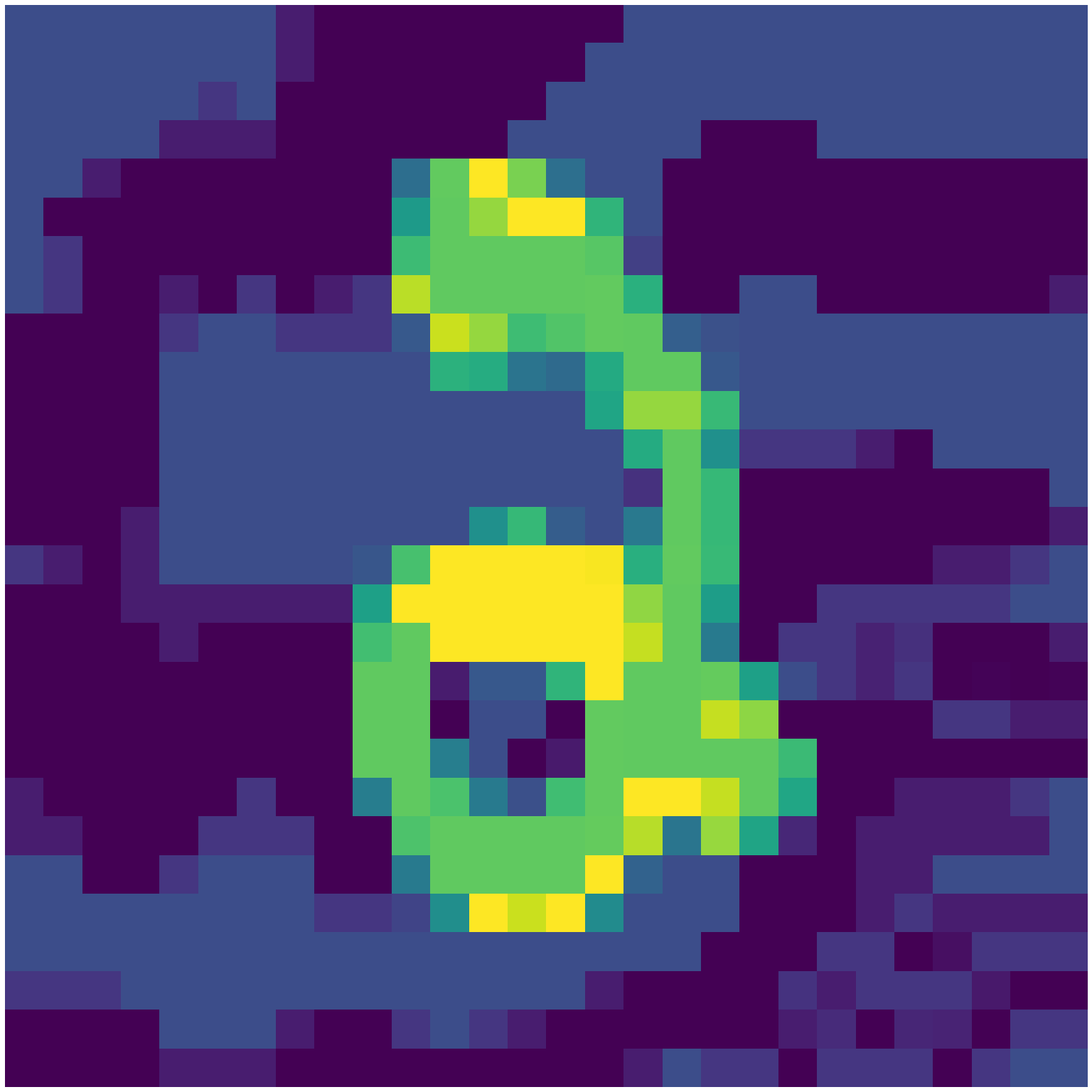}} 
 \\
  \parbox[c]{10em}{\includegraphics[trim=5 0 0 0,clip,width=9.5em]{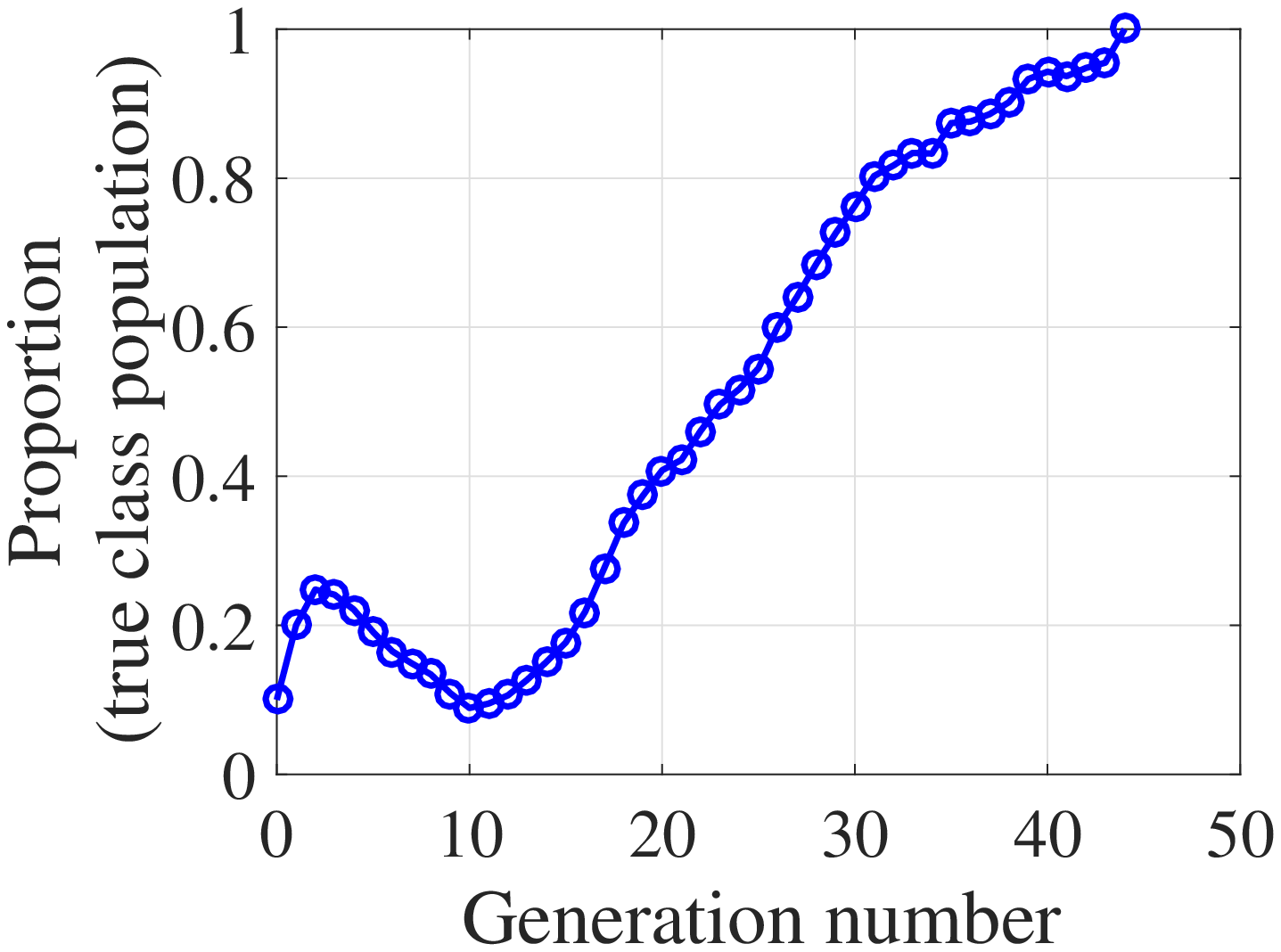}} 
  \\
 \parbox[c]{10em}{\includegraphics[trim=5 0 0 0,clip,width=9.5em]{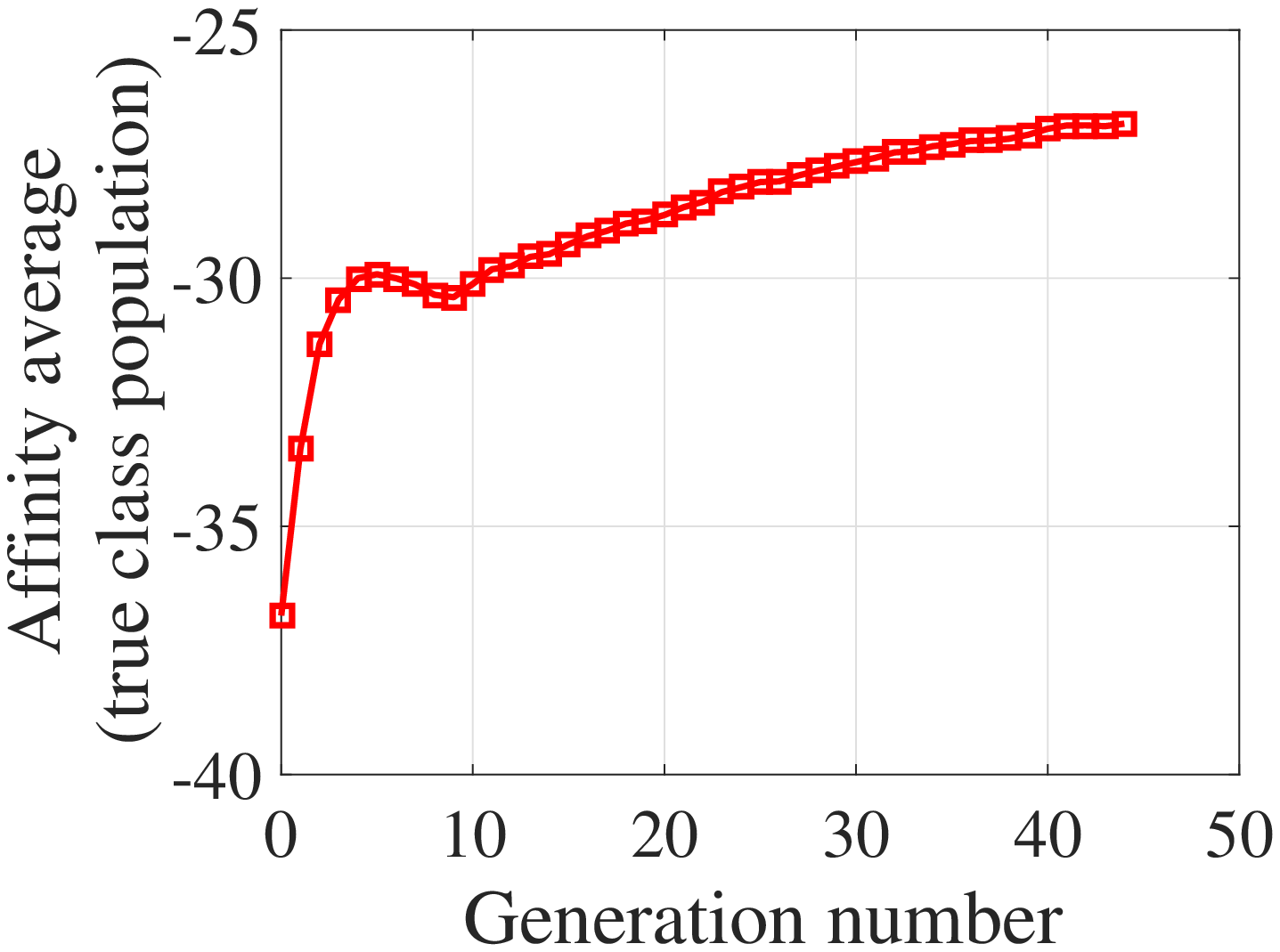}}
 \\
 \vspace{0.05in}
\begin{tabular}{@{\hskip 0.00in}c@{\hskip 0.00in}}
\parbox{9em}{\centering \scriptsize  \textbf{\begin{tabular}[c]{@{}c@{}}RAILS:2 VS DkNN:3\\RAILS confidenc: 1.0\\DkNN confidence: 0.6  \end{tabular}}} 
\end{tabular}

\end{tabular}

\end{tabular}
\end{tabular} 

\end{tabular}
  \end{adjustbox}
    \caption{Proportion and Average Affinity of True Class Population With Respect to Generation Number (RAILS on MNIST)}
  \label{fig: curves_running}
\end{figure}

\subsection{Visualization of RAILS Results}
Clonal expansion within RAILS creates new examples in each generation. To better understand the capability of RAILS, we can visualize the changes of some key indices during runtime. After the optimization step, the plasma data and memory data can be compared to the nearest neighbors found by the DkNN.

\paragraph{Visualization of RAILS Running Process.}
Figure~\ref{fig: curves_running} shows how the population and affinity score of the true class examples in each generation change when the generation number increases. We show two examples from MNIST here. DkNN only makes a correct prediction in the first example and gives a low confidence score for both examples. The first row depicts the proportion of the true class in each generation's population. Data from the true class occupies the majority of the population when the generation number increases, which indicates that RAILS can simultaneously obtain a correct prediction and a high confidence score. Meanwhile, clonal expansion over multiple generations produces increased affinity within the true class, as shown in the second row. Another observation is that RAILS requires fewer generations when DkNN makes a correct prediction, suggesting that affinity maturation occurs in fewer generations when the input is relatively easy to classify.

\paragraph{Visualization of RAILS Generated Examples.}

\begin{figure*}[h]
\centerline{\includegraphics[width=.78\textwidth]{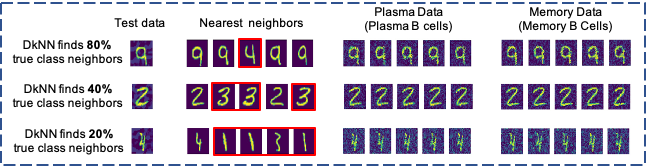}}
\vspace{-0.01in}
\caption{The Generated Plasma Examples and Memory Examples in a Single Layer}
\label{fig: plasma_memory}
\end{figure*}

Figure~\ref{fig: plasma_memory} shows the plasma data and memory data generated by RAILS. For the first example - digit $9$, DkNN gets $9$ in four out of five nearest neighbors. For the other two examples - digit $2$ and digit $1$, the nearest neighbors only contain a small amount of data from the true class. In contrast, the plasma data that RAILS generated are all from the true class, which provides correct prediction with confidence value 1. The memory data captures the information of the adversarial variants and is associated with the true label. They can be used to defend future adversarial inputs.

\subsection{Overall Performances on Different Scenarios}

We compare RAILS to CNN and DkNN in terms of SA and RA. DkNN uses $750$ samples from the training data as calibration data. RAILS leverages static learning to make the predictions. The results are shown in Table~\ref{tab2: acc_overall}. 
On MNIST with $\epsilon=60$, one can see that RAILS delivers a $5.62\%$ improvement in RA over DkNN without appreciable loss of SA. On CIFAR-10 (SVHN), RAILS leads to $4.74\%$ ($12.5\%$) and $20\%$ ($46\%$) robust accuracy improvements compared to DkNN and CNN, respectively. We also conduct experiments with Square Attack \citep*{andriushchenko2020square}, which is one of the black-box attacks. The results in Table~\ref{tab3: bba} show that RAILS improves the robust accuracy of DkNN by $4.7\%$ on CIFAR-10 with $\epsilon=24$. We refer readers to Section~3 in Supplementary for more results.

\begin{table}[h]
\vspace{-0.09in}
\begin{center}
\caption{
SA/RA Performance of RAILS versus CNN and DkNN}
\label{tab2: acc_overall}
\vspace{.04in}
\resizebox{0.45\textwidth}{!}{
\begin{tabular}{lc|c|c}
\hline
\hline 
& & SA & RA \\
\hline 
MNIST &\bf{RAILS (ours)} & 97.95\% & \bf{76.67\%}  \\

($\epsilon=60$) &CNN & \bf{99.16\%} & 1.01\% \\

& DkNN & 97.99\% & 71.05\%   \\
\hline 
SVHN  &\bf{RAILS (ours)}  & 90.62\% & \bf{48.26\%} \\

($\epsilon=8$) &CNN & \bf{94.55\%} & 1.66\% \\

&DkNN & 93.18\% & 35.7\% \\
\hline 
CIFAR-10 &\bf{RAILS (ours)} & 74\% & \bf{43\%}\\

($\epsilon=8$) &CNN & \bf{74.98\%} & 23.28\% \\

&DkNN & 74\% & 38.26\% \\

\hline
\hline
\end{tabular}}
\end{center}
\end{table}

\begin{table}[h]
\vspace{-0.16in}
\begin{center}
\caption{
SA/RA Performance of RAILS on CIFAR-10 under Square Attack \citep*{andriushchenko2020square}}
\label{tab3: bba}
\vspace{.04in}
\resizebox{0.48\textwidth}{!}{
\begin{tabular}{c||c|c|c}
\hline
\hline
&  SA& RA ($\epsilon=24$) & RA ($\epsilon=32$) \\
\hline 
\bf{RAILS (ours)}& 74\% & \bf{67.2\%} & \bf{57.2\%} \\
\hline
DkNN & 74\% & 62.5\% & 51.2\% \\
\hline
\hline
\end{tabular}}
\end{center}
\end{table}






\subsection{Single-Stage Adaptive Learning}

The previous experiments demonstrate that static learning is effective in predicting present adversarial inputs. 
Here we show that using single-stage adaptive learning (SSAL), i.e., one-time hardening in the test phase, can improve the robustness of depth classifier. We 
generate $3000$ memory data from a group of test data by feeding them into the RAILS framework. We then test whether the memory data can help DkNN defend the future evasion attack. The memory data works together with the original training data. We then randomly select another group of test data and generate $1000$ adversarial examples for evaluation. Table~\ref{tab4: acc_cnn_hard} shows that the SSAL improves RA of DkNN by $2.3\%$ with no SA loss using $3000$ memory data ($5\%$ of training data).



\begin{table}[h]
\vspace{-0.1in}
\begin{center}
\caption{
SA/RA Performance of DkNN (on $1000$ Adversarial Examples) Before and After Hardening the Classifier Through RAILS Single-Stage Adaptive Learning (SSAL)}
\label{tab4: acc_cnn_hard}
\vspace{.06in}
\resizebox{0.42\textwidth}{!}{
\begin{tabular}{l||c|c}
\hline
\hline 
 & SA & RA ($\epsilon=60$) \\
\hline
DkNN & 98.5\%  & 68.3\%  \\
\hline
\bf{DkNN-SSAL} & 98.5\%  & \bf{70.6\%}  \\

\hline
\hline
\end{tabular}}
\end{center}
\end{table}

\section{CONCLUSION}
Inspired by the immune system, we proposed a new defense framework for deep models. The proposed Robust Adversarial Immune-inspired Learning System (RAILS) has a one-to-one mapping to a simplified architecture immune system and its learning behavior aligns with {\em in vitro} biological experiments. RAILS incorporates static learning and adaptive learning, contributing to a robustification of predictions and dynamic model hardening, respectively. The experimental results demonstrate the effectiveness of RAILS. We believe this work is fundamental and delivers valuable principles for designing robust deep models. In future work, we will dig deeper into the mechanisms of the immune system's adaptive learning (life-long learning) and covariate shift adjustment, which will be  consolidated into our computational framework.

\bibliography{reference}




\newpage
\clearpage

\setcounter{section}{0}
\setcounter{figure}{0}
\makeatletter 
\renewcommand{\thefigure}{S\@arabic\c@figure}
\makeatother
\setcounter{table}{0}
\renewcommand{\thetable}{S\arabic{table}}

\setcounter{equation}{0}
\renewcommand{\theequation}{S\arabic{equation}}

\onecolumn
\section*{Supplementary Material}

\section{ADDITIONAL NOTES ON RAILS}
\paragraph{Computational complexity.} The RAILS's computational complexity is dominated by clonal expansion and affinity maturation. Given the length of the selected layers $N$, the population size $T$, and the maximum generation number $G$, the computational complexity is $O(TNG)$.

\paragraph{Early stopping criterion.} In Section 4.2 of the main text, we empirically show that RAILS requires fewer generations when DkNN makes a correct prediction. Therefore, $O(TNG)$ is the worst-case complexity. Considering the fast convergence of RAILS, one practical early stopping criterion is to check if a single class occupies most of the population, e.g., $100\%$.

\section{SETTINGS OF EXPERIMENTS}

\subsection{Datasets and Models}
We test RAILS on three public datasets: MNIST \citep*{lecun1998gradient}, SVHN \citep*{netzer2011reading}, and CIFAR-10 \citep*{Krizhevsky2009learning}. The MNIST dataset is a $10$-class handwritten digit database consisting of $60000$ training examples and $10000$ test examples. The SVHN dataset is another benchmark that is obtained from house numbers in Google Street View images. It contains $10$ classes of digits with $73257$ digits for training and $26032$ digits for testing. CIFAR-10 is a more complicated dataset that consists of $60000$ colour images in $10$ classes. There are $50000$ training images and $10000$ test images. We use a four-convolutional-layer neural network for MNIST, and VGG16 \citep*{SZ2014} for SVHN and CIFAR-10. For MNIST and SVHN, we conduct the clonal expansion in the inputs. To provide better feature representations for CIFAR-10, we use an adversarially trained model on $\epsilon=2$ and conduct the clonal expansion in a shallow layer.

\subsection{Threat Models}
Though out this paper, we consider three different types of attacks: (1) Projected Gradient Descent (PGD) attack \citep*{madry17} - We implement $20$-step PGD attack for MNIST, and $10$-step PGD attack for SVHN and CIFAR-10. The attack strength is $\epsilon = 40/60/76.5$ for MNIST, and $\epsilon = 4/8$ for SVHN and CIFAR-10. (2) Fast Gradient Sign Method \citep*{goodfellow2014explaining} - The attack strength is $\epsilon = 76.5$ for MNIST, and $\epsilon = 4/8$ for SVHN and CIFAR-10. (3) Square Attack \citep*{andriushchenko2020square} - We implement $50$-step attack for MNIST, and $30$-step attack for CIFAR-10. The attack strength is $\epsilon = 76.5$ for MNIST, and $\epsilon = 24/32$ for CIFAR-10.

\subsection{Parameter Selection}
By default, we set the size of the population $T=1000$, the maximum number of generations $G=50$, and the mutation probability $\rho=0.15$. To speed up the algorithm, we stop when the newly generated examples are all from the same class. For MNIST, we set the mutation range parameters $\delta_{\text{min}}=0.05 (12.75), \delta_{\text{max}}=0.15 (38.25)$. The sampling temperature $\tau$ in each layer is set to $3, 18, 18,$ and $72$. The principle of selecting $\tau$ is to make sure that the high affinity samples do not dominate the whole dataset, i.e. the algorithm will not stop after the first generation. In general, we assign larger values to hidden layers with
smaller length. We find that our method works well in a wide range of $\tau$. Similarly, we set $\tau=300$ for CIFAR-10 and SVHN. Considering CIFAR-10 and SVHN are more sensitive to small perturbations, we set the mutation range parameters $\delta_{\text{min}}=0.005 (1.275), \delta_{\text{max}}=0.015 (3.825)$.

\section{ADDITIONAL EXPERIMENTS}

\subsection{Additional Comparisons on MNIST}

\begin{figure}[h]
\vspace*{-0.1in}
  \centering
  \begin{adjustbox}{max width=0.42\textwidth }
  \begin{tabular}{@{\hskip 0.00in}c  @{\hskip 0.00in}c @{\hskip 0.02in} @{\hskip 0.02in} c @{\hskip 0.02in} @{\hskip 0.02in}c }
 \begin{tabular}{@{}c@{}}  
\vspace*{0.005in}\\
\rotatebox{90}{\parbox{17em}{\centering \Large \textbf{Adv examples ($\epsilon=60$)}}}
 \\
\rotatebox{90}{\parbox{20em}{\centering \Large \textbf{Clean examples}}}
\\
\end{tabular} 
&
\begin{tabular}{@{\hskip 0.02in}c@{\hskip 0.02in}}
     \begin{tabular}{@{\hskip 0.00in}c@{\hskip 0.00in}}
     \parbox{10em}{\centering \Large Conv1}  
    \end{tabular} 
    \\
 \parbox[c]{20em}{\includegraphics[width=20em]{Figs/conv1eps60new}} 
 \\
 \parbox[c]{20em}{\includegraphics[width=20em]{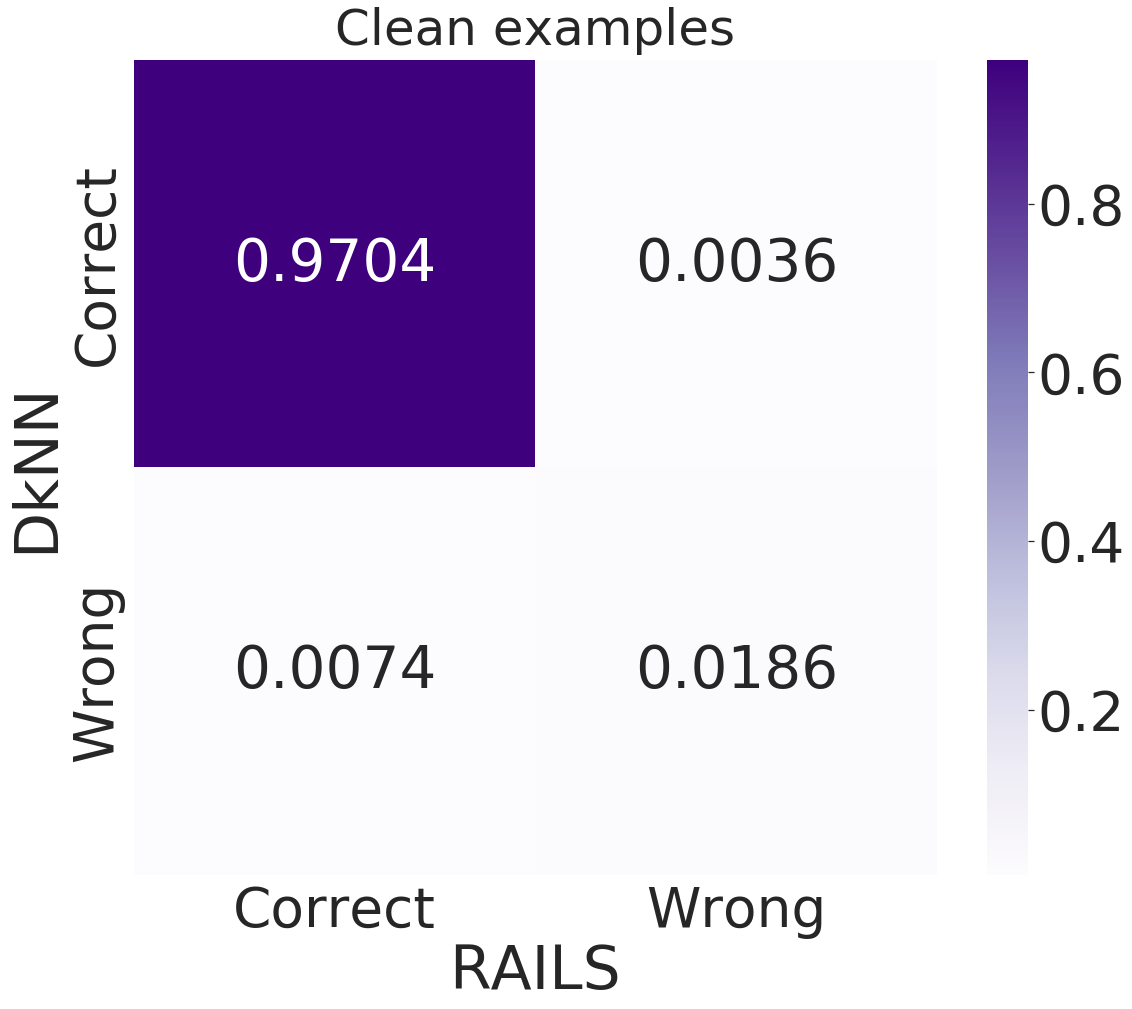}}
\end{tabular}
&
 \begin{tabular}{@{\hskip 0.02in}c@{\hskip 0.02in}}
      \begin{tabular}{@{\hskip 0.00in}c@{\hskip 0.00in}}
     \parbox{10em}{\centering \Large Conv2}
    \end{tabular} 
    \\
 \parbox[c]{20em}{\includegraphics[width=20em]{Figs/conv2eps60new}} 
 \\
 \parbox[c]{20em}{\includegraphics[width=20em]{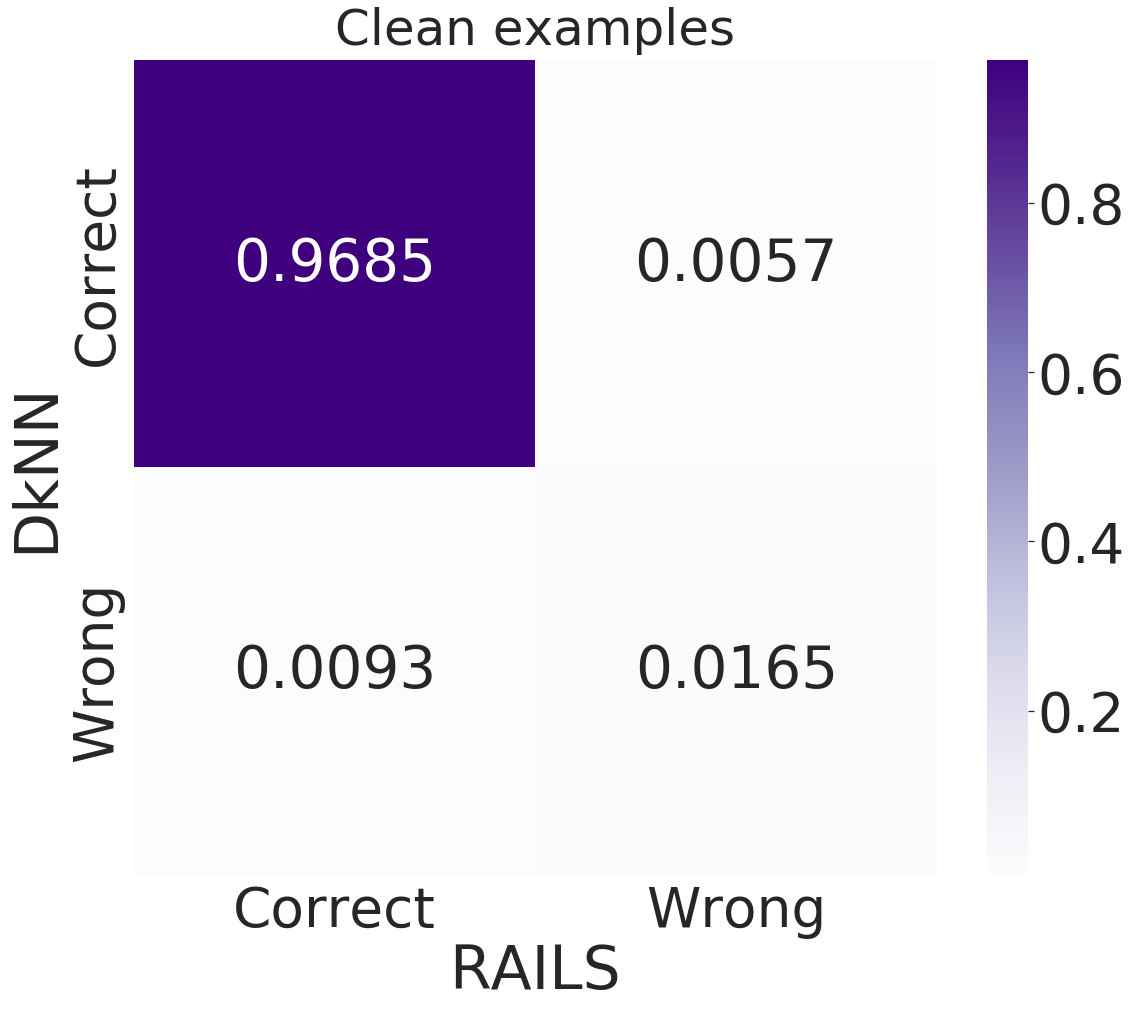}}
\end{tabular}

\end{tabular}
  \end{adjustbox}
    \caption{Confusion Matrices in Convolutional Layer 1 and 2 (RAILS vs. DkNN - $\epsilon=60$)}
  \label{figS: conf_mat}
\end{figure}

Figure~\ref{figS: conf_mat} provides the confusion matrices for clean examples classifications and adversarial examples classifications in Conv1 and Conv2 when $\epsilon=60$. The confusion matrices in Figure~\ref{figS: conf_mat} show that RAILS has fewer incorrect predictions for those data that DkNN gets wrong. Each value in Figure~\ref{figS: conf_mat} represents the percentage of intersections of RAILS (correct or wrong) and DkNN (correct or wrong).

\begin{figure}[h]
\vspace{-0.03in}
\centering
\begin{minipage}{0.3\linewidth}
\centerline{
\includegraphics[width=.84\linewidth]{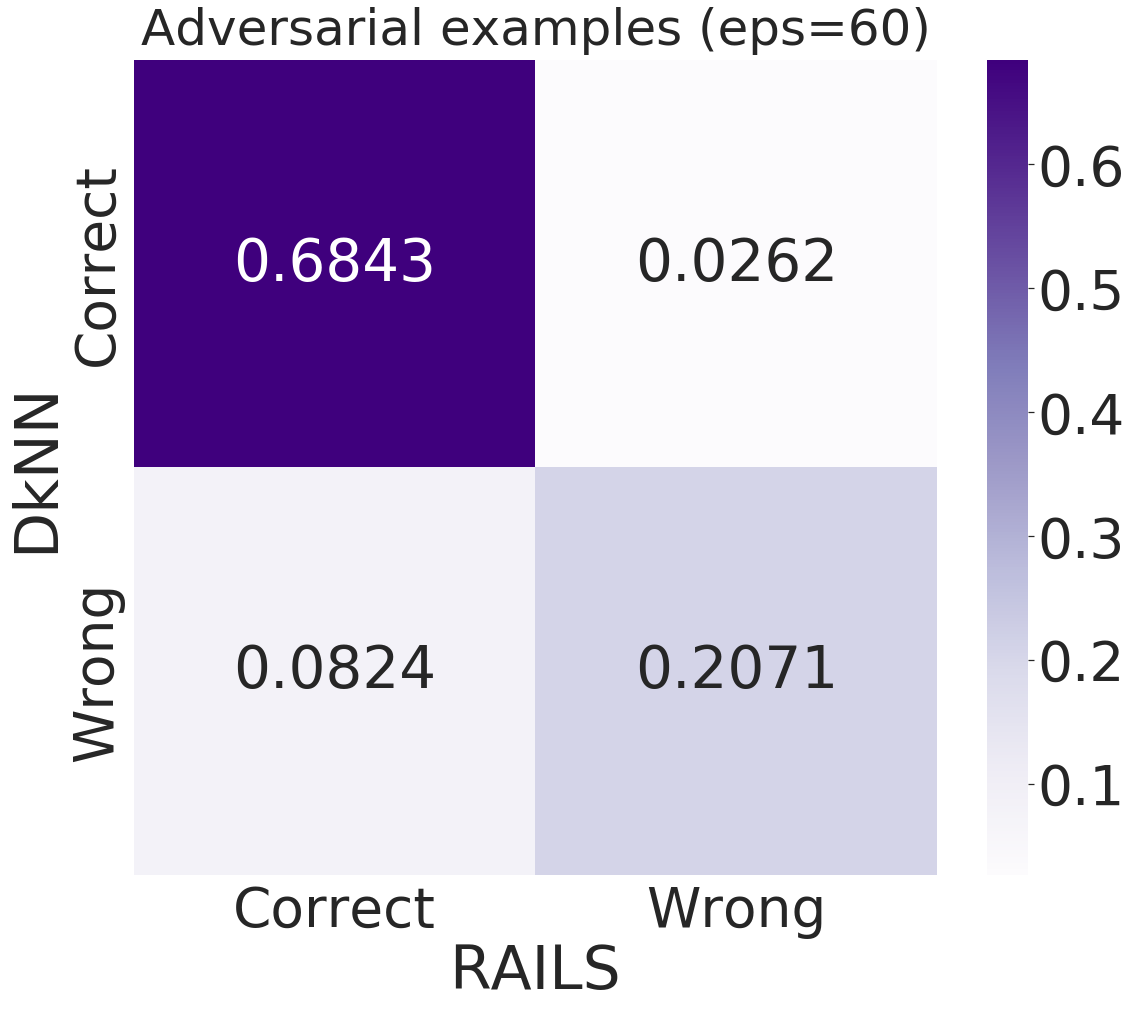}}
\end{minipage}
\centering
\begin{minipage}{0.3\linewidth}
\centerline{
\includegraphics[width=.84\linewidth]{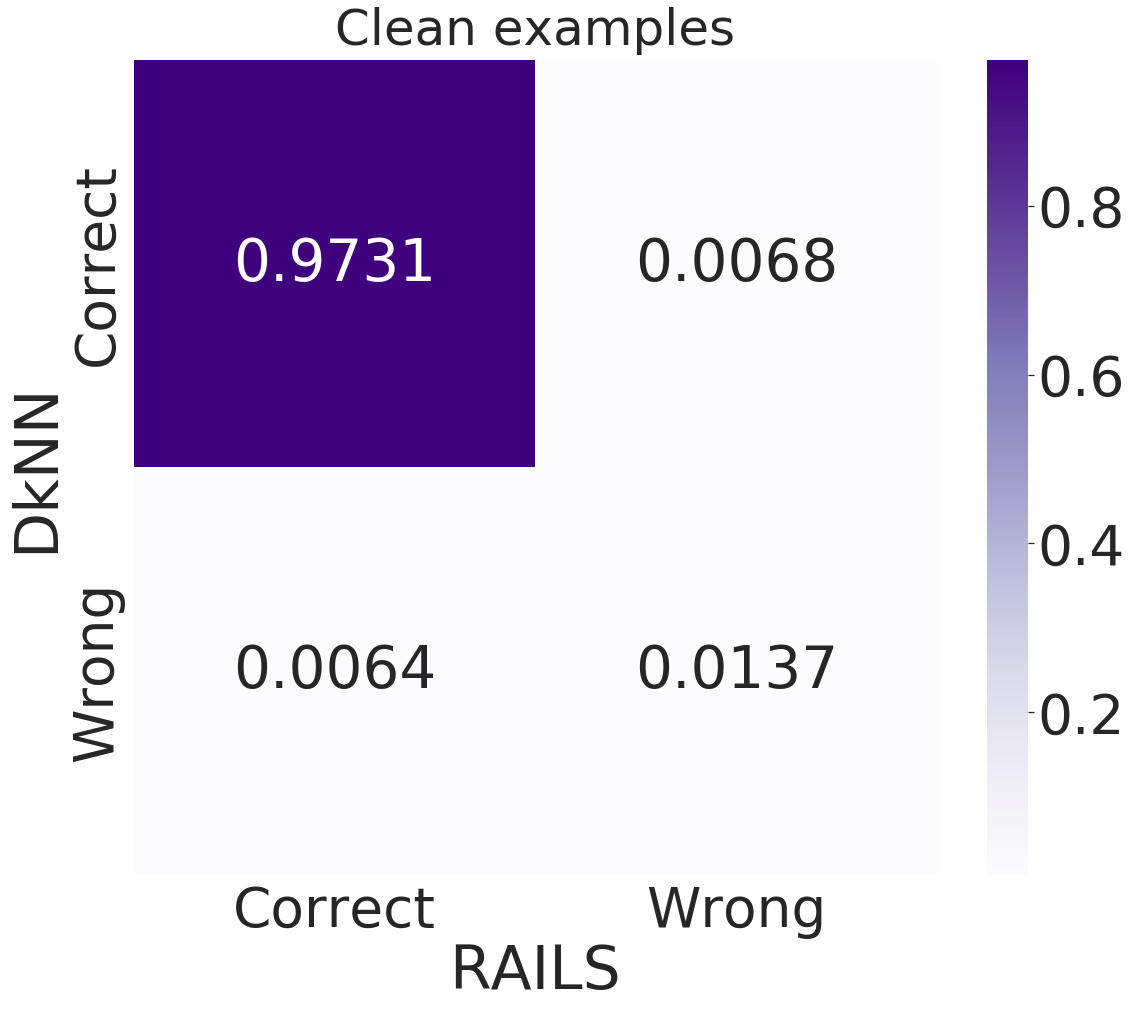}}
\end{minipage}
\caption{Confusion Matrices (RAILS vs. DkNN - $\epsilon=60$)}
\label{fig: conf_mat_overall}
\end{figure}

Figure~\ref{fig: conf_mat_overall} shows the confusion matrices of the overall performance when $\epsilon=60$. The confusion matrices indicate that RAILS' correct predictions agree with a majority of DkNN's correct predictions and disagree with DkNN's wrong predictions.

We also show the SA/RA performance of RAILS under PGD attack and FGSM when $\epsilon=76.5$. The results in Table~\ref{tabS: mnist_pgd_fgsm} indicate that RAILS can reach higher RA than DkNN with close SA.

\begin{table}[h]
\begin{center}
\caption{
SA/RA Performance of RAILS on MNIST under PGD Attack and FGSM ($\epsilon=76.5$)}
\label{tabS: mnist_pgd_fgsm}
\vspace{.04in}
\resizebox{0.55\textwidth}{!}{
\begin{tabular}{c||c|c|c}
\hline
\hline
&  SA &  RA (PGD) & RA (FGSM)\\
\hline 
\bf{RAILS (ours)}& 97.95\% & \bf{58.62\%}  & \bf{61.67\%} \\
\hline
DkNN& 97.99\%  & 47.05\% & 52.23\%\\
\hline
\hline
\end{tabular}}
\end{center}
\end{table}

In Table~\ref{tabS: bba}, we provide the experimental results with Square Attack  \citep*{andriushchenko2020square} (a black-box attack) showing that RAILS improves the robust accuracy of DkNN by $1.35\%$ ($11\%$ attack success rate) on MNIST with $\epsilon=76.5$.

\begin{table}[h]
\begin{center}
\caption{
SA/RA Performance of RAILS on MNIST under Square Attack with $\epsilon=76.5$}
\label{tabS: bba}
\vspace{.04in}
\resizebox{0.4\textwidth}{!}{
\begin{tabular}{c||c|c}
\hline
\hline
&  SA&  RA\\
\hline 
\bf{RAILS (ours)}& 97.95\% & \bf{89.35\%}  \\
\hline
DkNN& 97.99\%  & 88\% \\
\hline
\hline
\end{tabular}}
\end{center}
\end{table}

\subsection{Additional Comparisons on CIFAR-10}
In this subsection, we test RAILS on CIFAR-10 under PGD attack and FGSM with attack strength $\epsilon=4/8$. The results are shown in Figure~\ref{tabS: cifar} and Figure~\ref{tabS: cifar_fgsm}. RAILS outperforms DkNN and CNN on different attack types and strengths. We also find that the difference of RA between RAILS and DkNN increases when $\epsilon$ increases, indicating that RAILS can defend stronger attacks.

\begin{table}[h]
\begin{center}
\caption{
SA/RA Performance of RAILS on CIFAR-10 under PGD Attack}
\label{tabS: cifar}
\vspace{.06in}
\resizebox{0.52\textwidth}{!}{
\begin{tabular}{l||c|c|c}
\hline
\hline 
 & SA & RA ($\epsilon=4$) & RA ($\epsilon=8$) \\
\hline 
\bf{RAILS (ours)} & 74\% & \bf{58\%} & \bf{43\%}  \\
\hline
CNN & \bf{74.98\%} & 47.15\% & 23.28\% \\
\hline
DkNN & 74\% & 54\% & 38.26\% \\

\hline
\hline
\end{tabular}}
\end{center}
\end{table}

\begin{table}[h]
\begin{center}
\caption{
SA/RA Performance of RAILS on CIFAR-10 under FGSM}
\label{tabS: cifar_fgsm}
\vspace{.06in}
\resizebox{0.5\textwidth}{!}{
\begin{tabular}{l||c|c|c}
\hline
\hline 
 & SA & RA ($\epsilon=4$) & RA ($\epsilon=8$)  \\
\hline 
\bf{RAILS (ours)} & 74\% & \bf{60\%} & \bf{46.2\%} \\
\hline
CNN & \bf{74.98\%} & 50.55\% & 31.19\% \\
\hline
DkNN & 74\% & 57.5\% & 42.3\% \\

\hline
\hline
\end{tabular}}
\end{center}
\end{table}

\end{document}